# Learning Max-Margin Tree Predictors

Ofer Meshi[†*]    Elad Eban[†*]    Gal Elidan[‡]    Amir Globerson[†]

[†] School of Computer Science and Engineering
[‡] Department of Statistics
The Hebrew University of Jerusalem, Israel

## Abstract

Structured prediction is a powerful framework for coping with joint prediction of interacting outputs. A central difficulty in using this framework is that often the correct label dependence structure is unknown. At the same time, we would like to avoid an overly complex structure that will lead to intractable prediction. In this work we address the challenge of learning tree structured predictive models that achieve high accuracy while at the same time facilitate efficient (linear time) inference. We start by proving that this task is in general NP-hard, and then suggest an approximate alternative. Our CRANK approach relies on a novel Circuit-RANK regularizer that penalizes non-tree structures and can be optimized using a convex-concave procedure. We demonstrate the effectiveness of our approach on several domains and show that its accuracy matches that of fully connected models, while performing prediction substantially faster.

## 1 Introduction

Numerous applications involve joint prediction of complex outputs. For example, in document classification the goal is to assign the most relevant (possibly multiple) topics to each document; in gene annotation, we would like to assign each gene a set of relevant functional tags out of a large set of possible cellular functions; in medical diagnosis, we would like to identify all the diseases a given patient suffers from. Although the output space in such problems is typically very large, it often has intrinsic *structure* which can be exploited to construct efficient predictors. Indeed, in recent years using *structured output prediction* has resulted in state-of-the-art results in many real-worlds problems from computer vision, natural language processing, computational biology, and other fields [Bakir et al., 2007]. Such predictors can be learned from data using formulations such as Max-Margin Markov Networks ($M^3N$) [Taskar et al., 2003, Tsochantaridis et al., 2006], or conditional random fields (CRF) [Lafferty et al., 2001].

While the prediction and the learning tasks are generally computationally intractable [Shimony, 1994, Sontag et al., 2010], for some models they can be carried out efficiently. For example, when the model consists of pairwise dependencies between output variables, and these form a tree structure, prediction can be computed efficiently using dynamic programming at a linear cost in the number of output variables [Pearl, 1988]. Moreover, despite their simplicity, tree structured models are often sufficiently expressive to yield highly accurate predictors. Accordingly, much of the research on structured prediction focused on this setting [e.g., Lafferty et al., 2001, Collins, 2002, Taskar et al., 2003, Tsochantaridis et al., 2006].

Given the above success of tree structured models, it is unfortunate that in many scenarios, such as a document classification task, there is no obvious way in which to choose the most beneficial tree. Thus, a natural question is how to find the tree model that best fits a given structured prediction problem. This is precisely the problem we address in the current paper. Specifically, we ask what is the tree structure that is optimal in terms of a max-margin objective [Taskar et al., 2003]. Somewhat surprisingly, this optimal tree problem has received very little attention in the context of discriminative structured prediction (the most relevant work is Bradley and Guestrin [2010] which we address in Section 6).

Our contributions are as follows. We begin by proving that it is NP-hard in general to find the optimal max-margin predictive tree, in marked contrast to the generative case where the optimal tree can be learned efficiently [Chow and Liu, 1968]. To cope with this theoretical barrier, we propose an approximation scheme that uses regularization to penalize non-tree models. Concretely, we propose a regularizer that is based on the *circuit rank* of a graph [Berge, 1962], namely the minimal number of edges that need to be removed from the graph in order to obtain a tree. Minimization of the resulting objective is still difficult, and we further approximate it using a difference of continuous convex envelopes. The resulting objective can then be readily optimized using the convex concave procedure [Yuille and Rangarajan, 2003].

---

[*]Authors contributed equally.

We apply our method to synthetic and varied real-world structured output prediction tasks. First, we show that the learned tree model is competitive with a fully connected max-margin model that is substantially more computationally demanding at prediction time. Second, we show that our approach is superior to several baseline alternatives (e.g., greedy structure learning) in terms of generalization performance and running time.

## 2 The Max-margin Tree

Let $x$ be an input vector (e.g., a document) and $y$ a discrete output vector (e.g., topics assigned to the document, where $y_i = 1$ when topic $i$ is addressed in $x$). As in most structured prediction approaches, we assume that inputs are mapped to outputs according to a linear discrimination rule: $y(x; w) = \text{argmax}_{y'} w^\top \phi(x, y')$, where $\phi(x, y)$ is a function that maps input-output pairs to a feature vector, and $w$ is the corresponding weight vector. We will call $w^\top \phi(x, y')$ the *score* that is assigned to the prediction $y'$ given an input $x$.

Assume we have a set of $M$ labeled pairs $\{(x^m, y^m)\}_{m=1}^M$, and would like to learn $w$. In the $M^3N$ formulation proposed by Taskar et al. [2003], $w$ is learned by minimizing the following (regularized) structured hinge loss:

$$\ell(w) = \frac{\lambda}{2}\|w\|^2 + \frac{1}{M}\sum_m h^m(w),$$

where

$$h^m(w) = \max_y \left[ w^\top \phi(x^m, y) + \Delta(y, y^m) \right] - w^\top \phi(x^m, y^m), \quad (1)$$

and $\Delta(y, y^m)$ is a label-loss function measuring the cost of predicting $y$ when the true label is $y^m$ (e.g., 0/1 or Hamming distance). Thus, the learning problem involves a loss-augmented prediction problem for each training example.

Since the space of possible outputs may be quite large, maximization of $y$ can be computationally intractable. It is therefore useful to consider score functions that decompose into simpler ones. One such decomposition that is commonly used consists of scores over single variables and pairs of variables that correspond to nodes and edges of a graph $G$, respectively:

$$w^\top \phi(x, y) = \sum_{ij \in E(G)} w_{ij}^\top \phi_{ij}(x, y_i, y_j) + \sum_{i \in V(G)} w_i^\top \phi_i(x, y_i). \quad (2)$$

Importantly, when the graph $G$ has a tree structure then the maximization over $y$ can be solved exactly and efficiently using dynamic programming algorithms (e.g., Belief Propagation [Pearl, 1988]).

As mentioned above, we consider problems where there is no natural way to choose a particular tree structure, and our goal is to learn the optimal tree from training data. We next formalize this objective.

In a tree structured model, the set of edges $ij$ in Eq. (2) forms a tree. This is equivalent to requiring that the vectors $w_{ij}$ in Eq. (2) be non-zero only on edges of some tree. To make this precise, we first define, for a *given* spanning tree $T$, the set $\mathcal{W}_T$ of weight vectors that "agree" with $T$:[1]

$$\mathcal{W}_T = \{w : ij \notin T \implies w_{ij} = 0\}. \quad (3)$$

Next we consider the set $\mathcal{W}_\cup$ of weight vectors that agree with *some* spanning tree. Denote the set of all spanning trees by $\mathcal{T}$, then: $\mathcal{W}_\cup = \bigcup_{T \in \mathcal{T}} \mathcal{W}_T$. The problem of finding the optimal max-margin tree predictor is therefore:

$$\min_{w \in \mathcal{W}_\cup} \ell(w). \quad (4)$$

We denote this as the $M^{Tree}N$ problem. In what follows, we first show that this problem is NP-hard, and then present an approximation scheme.

## 3 Learning $M^3N$ Trees is NP-hard

We start by showing that learning the optimal tree in the discriminative max-margin setting is NP-hard. As noted, this is somewhat of a surprise given that the best tree is easily learned in the generative setting [Chow and Liu, 1968], and that tree structured models are often used due to their computational advantages.

In particular, we consider the problem of deciding whether there exists a tree structured model that correctly labels a given dataset (i.e., deciding whether the dataset is separable with a tree model). Formally, we define the $M^{Tree}N$ *decision problem* as determining whether the following set is empty:

$$\left\{w \in \mathcal{W}_{\cup} \middle| w^\top \phi(x^m, y^m) \geq w^\top \phi(x^m, y) + \Delta(y, y^m) \; \forall m, y\right\}. \quad (5)$$

---

[1]Note that weights corresponding to single node features are not restricted.

To facilitate the identifiability of the model parameters that is later needed, we adopt the formalism of Sontag et al. [2010] and define the score:

$$\begin{aligned}
S(y; x, T, w) &= \sum_{ij \in T} w_{ij}^\top \phi_{ij}(x, y_i, y_j) + \sum_i (w_i^\top \phi_i(x, y_i) + x_i(y_i)) \\
&\equiv \sum_{ij \in T} \theta_{ij}(y_i, y_j) + \sum_i \theta_i(y_i) + \sum_i x_i(y_i), \quad (6)
\end{aligned}$$

where $x_i(y_i)$ is a bias term which does not depend on $w$,[2] and for notational convenience we have dropped the dependence of $\theta$ on $x$ and $w$. We can now reformulate the set in Eq. (5) as:

$$\left\{T, w \middle| S(y^m; x^m, T, w) \geq \max_y S(y; x^m, T, w) \; \forall m\right\}, \quad (7)$$

where, for simplicity, we omit the label loss $\Delta(y, y^m)$. This is valid since the bias terms already ensure that the trivial solution $w = 0$ is avoided. With this reformulation, we can now state the hardness result.

**Theorem 3.1.** *Given a set of training examples $\{(x^m, y^m)\}_{m=1}^M$, it is NP-hard to decide whether there exists a tree $T$ and weights $w$ such that $\forall m, y^m = \text{argmax}_y S(y; x^m, T, w)$.*

*Proof.* We show a reduction from the NP-hard *bounded-degree spanning tree* (BDST) problem to the $M^{Tree}N$ decision problem defined in Eq. (7).[3] In the BDST problem, given an undirected graph $G$ and an integer $D$, the goal is to decide whether there exists a spanning tree with maximum degree $D$ (for $D = 2$ this is the Hamiltonian path problem, hence the NP-hardness).

Given an instance of BDST we construct an instance of problem Eq. (7) on the same graph $G$ as follows. First, we define variables $y_1, \ldots, y_n$ that take values in $\{0, 1, \ldots, n\}$, where $n = |V(G)|$. Second, we will define the parameters $\theta_i(y_i)$ and $\theta_{ij}(y_i, y_j)$ and bias terms $x_i(y_i)$ in such a way that solving the $M^{Tree}N$ decision problem will also solve the BDST problem. To complement this, we will define a set of training examples which are separable only by the desired parameters. For clarity of exposition, we defer the proof that these parameters are identifiable using a polynomial number of training examples to App. A.

The singleton parameters are

$$\theta_i(y_i) = \begin{cases} D & i = 1, y_1 = 0 \\ 0 & \text{otherwise,} \end{cases} \quad (8)$$

and the pairwise parameters for $ij \in E(G)$ are:

$$\theta_{ij}(y_i, y_j) = \begin{cases} -n^2 & y_i \neq y_j \\ 0 & y_i = y_j = 0 \\ 1 & y_i = y_j = i \\ 1 & y_i = y_j = j \\ 0 & \text{otherwise.} \end{cases} \quad (9)$$

Now consider the case where the bias term $x_i(y_i)$ is identically zero. In this case the score for (non-zero) uniform assignments equals the degree of the vertex in $T$. That is, $S(i, \ldots, i; 0, T, \theta) = \deg_T(i)$ since $\theta_{ij}(i, i) = 1$ for all $j \in N(i)$ and the other parameter values are zero. The score for the assignment $y = 0$ is $S(0, \ldots, 0; 0, T, \theta) = D$, and for non-uniform assignments we get a negative score $S(y; 0, T, \theta) < 0$. Therefore, the maximization over $y$ in Eq. (7) reduces to a maximization over $n$ uniform states (each corresponding to a vertex in the graph). The maximum value is thus the maximum between $D$ and the maximum degree in $T$: $\max_y S(y; 0, T, \theta) = \max\{D, \max_i \deg_T(i)\}$.

It follows that, if we augment the training set that realizes the above parameters (see App. A) with a single training example where $x^m = y^m = (0, \ldots, 0)$, the set of Eq. (7) can now be written as:

$$\left\{T \middle| D \geq \max_i \deg_T(i)\right\}.$$

Thus, we have that the learning problem is separable if and only if there exists a bounded degree spanning tree in $G$. This concludes the reduction, and we have shown that the decision problem in Theorem 3.1 is indeed NP-hard. □

The above hardness proof illustrates a striking difference between generative learning of tree models (i.e., Chow Liu) and discriminative learning (our NP-hard setting). Clearly, we do not want to abandon the discriminative setting and trees remain computationally appealing at test time. Thus, in the rest of the paper, we propose practical approaches for learning a tree structured predictive model and demonstrate empirically that our method is competitive.

---
[2] As usual, the bias term can be removed by fixing some elements of $w$ to 1.

[3] A related reduction is given in Aissi et al. [2005], Theorem 8.

## 4 Tree-Inducing Regularization

Due to the hardness of the tree learning problem, we next develop an approximation scheme for it. We begin with the exact formulation of the problem, and then introduce an approximate formulation along with an optimization procedure. Our construction relies on several properties of submodular functions and their convex envelopes.

### 4.1 Exact Tree Regularization

As described in Section 2, we would like to find a tree structured weight vector $w \in \mathcal{W}_\cup$ that minimizes the empirical hinge loss. The main difficulty in doing so is that the sparsity pattern of $w$ needs to obey a fairly complex constraint, namely being tree structured. This is in marked contrast to popular sparsity constraints such as an upper bound on the number of non-zero values, a constraint that does not take into account the resulting global structure.

To overcome this difficulty, we will formulate the exact learning problem via an appropriate regularization. We begin by defining a function that maps $w$ to the space of edges: $\pi : \mathbb{R}^d \mapsto \mathbb{R}^{|E|}$, where $E$ corresponds to all edges of the full graph. Specifically, the component in $\pi$ corresponding to the edge $ij$ is:

$$\pi_{ij}(w) = \|w_{ij}\|_1.$$

Now, denote by $Supp(\pi(w))$ the set of coordinates in $\pi(w)$ that are non-zero. We would like the edges corresponding to these coordinates to form a tree graph. Thus, we wish to define a set function $F(Supp(\pi(w)))$ which will be equal to zero if $Supp(\pi(w))$ conforms to *some* tree structure, and a positive value otherwise. If we then add $\beta F(Supp(\pi(w)))$ to the objective in Eq. (4) with $\beta$ large enough, the resulting $w$ will be tree structured. The optimization problem is then:

$$\min_w \ell(w) + \beta F(Supp(\pi(w))). \qquad (10)$$

In what follows, we define a set function $F(A)$ with the desired properties. That is, we seek a function that takes a set of edges $A$ and outputs zero if they correspond to a tree, and a positive number otherwise. Intuitively, it is also desirable to define the function such that its value increases as the graph becomes less "tree-like". To make this concrete we define a measure for "treeness" as the minimum number of edges that need to be removed from $A$ in order to reduce it to a tree structure. This measure is also known as the *circuit-rank* of the graph [Berge, 1962]. Formally:

$$r = |A| + c(A) - n,$$

where $c(A)$ is the number of connected components in the graph, and $n$ is the number of vertices. We note that the circuit rank is also the co-rank of the graphic matroid, and is hence supermodular.

Putting it all together, we have that our desired tree-inducing function is given by:

$$F(Supp(\pi(w))) = |Supp(\pi(w))| + c(Supp(\pi(w))) - n.$$

Of course, given our hardness result, optimizing Eq. (10) with the above $F(A)$ is still computationally hard. From an optimization perspective, the difficulty comes from the non-convexity of the the above function in $w$. Optimization is further complicated by the fact that it is highly non-smooth, similarly to the $\ell_0$ norm. In the next section we suggest a smoothed approximation that is more amenable to optimization.

We also note that in Eq. (10) $w$ is generally not tree structured, so the maximization over $y$ in Eq. (1) is not necessarily tractable. Therefore, we replace the hinge loss in Eq. (1) with its *overgenerating* approximation [Finley and Joachims, 2008], known as linear programming (LP) relaxation [e.g., see Meshi et al., 2010]. This is achieved by formulating the optimization in Eq. (1) as an integer LP and then relaxing the integrality requirement, allowing fractional solutions. Importantly, for tree structured graphs this approximation is in fact exact [Wainwright and Jordan, 2008]. This implies that if there exists a tree model that separates the data, it will be found even when using this relaxation in Eq. (10). With slight abuse of notation, we keep referring to the approximate objective as $\ell(w)$.

### 4.2 Approximate Tree Regularization

The function $F(A)$ is a set function applied to the support of $\pi(w)$. Bach [2010] (Proposition 1) shows that when $F$ is submodular *and* non-decreasing, the convex envelope of $F(Supp(\pi(w)))$ can be calculated efficiently. This is desirable since the convex envelope then serves as a convex regularizer. Furthermore, this convex envelope can be elegantly understood as the Lovász extension $f$ of $F$, applied to $|\pi(w)|$ (in our case, $|\pi(w)| = \pi(w)$). Unfortunately, the circuit-rank $r$ does not satisfy these conditions, since it is supermodular and non-decreasing (in fact, its convex envelope is a constant).

To overcome this difficulty, we observe that $F(A)$ can be decomposed in a way that allows us to use the result of Bach [2010]. Specifically, we can write $F(A) = F_1(A) - F_2(A)$, where

$$F_1(A) = |A| \;, \; F_2(A) = n - c(A). \qquad (11)$$

$F_1(A)$ is simply the cardinality function which is modular and increasing. Furthermore, $F_2(A)$ is the rank of the graphic matroid [Oxley, 2006], and is hence submodular. It is also easy to see that $F_2(A)$ is non-decreasing. Thus, both functions satisfy the conditions of Proposition 1 in Bach [2010] and their convex envelopes can be found in closed form, as characterized in the following corollaries.

**Corollary 4.1.** *The convex envelope of $F_1(Supp(\pi(w)))$ is $f_1(\pi(w)) = \sum_{ij} \pi_{ij}(w) = \|w\|_1$.*

*Proof.* Follows directly from Prop. 1 in Bach [2010] and the fact that the Lovász extension of the cardinality function is the $\ell_1$ norm. □

**Corollary 4.2.** *The convex envelope of $F_2(Supp(\pi(w)))$ is $f_2(\pi(w))$, defined as follows. Sort the elements of $\pi(w)$ in decreasing order, and construct a maximum-spanning-tree with this ordering as in Kruskal's algorithm [Kruskal, 1956]. Denoting the resulting tree by $T(\pi(w))$, we obtain*

$$f_2(\pi(w)) = \sum_{ij \in T(\pi(w))} \pi_{ij}(w) = \sum_{ij \in T(\pi(w))} \|w_{ij}\|_1$$

*Proof.* Let $(ij)_k$ denote the $k^{th}$ edge when sorting $\pi(w)$, then the Lovász extension $f_2$ of $F_2$ at $\pi(w)$ is:

$$\sum_{k=1}^{|E|} \pi_{(ij)_k}(w)[F_2(\{(ij)_1,\ldots,(ij)_k\}) - F_2(\{(ij)_1,\ldots,(ij)_{k-1}\})]$$
$$= \sum_{ij \in T(\pi(w))} \pi_{ij}(w),$$

where we have used Eq. (11) and the fact that the number of connected components decreases by one only when introducing edges in Kruskal's tree. The desired result follows from Prop. 1 in [Bach, 2010]. □

We now approximate $F(Supp(\pi(w)))$ as a difference of the two corresponding convex envelopes, denoted by $f(\pi(w))$:

$$f(\pi(w)) \equiv f_1(\pi(w)) - f_2(\pi(w)) = \sum_{ij \notin T(\pi(w))} \|w_{ij}\|_1 \quad (12)$$

This function has two properties that make it computationally and conceptually appealing:

- $f(\pi(w))$ is a difference of two convex functions so that a local minimum can be easily found using the convex concave procedure (see Section 4.3).

- The set $\{ij \notin T(\pi(w))\}$ are precisely these edges that form a cycle when added according to the order implied by $\pi(w)$. Thus, the penalty we use corresponds to the magnitude of $\|w_{ij}\|_1$ on the edges that form cycles, namely the non-tree edges.

### 4.3 Optimizing with Approximate Tree Regularization

Using the tree inducing regularizer from the previous section, our overall optimization problem becomes:

$$\min_w \ell(w) + \beta f_1(\pi(w)) - \beta f_2(\pi(w)).$$

Since the function $f_2(\pi(w))$ is rather elaborate, optimizing the above objective still requires care. In what follows, we introduce a simple procedure for doing so that utilizes the convex concave procedure (CCCP) [Yuille and Rangarajan, 2003].[4] Recall that CCCP is applicable for an objective function (to be minimized) that is a sum of a convex and concave functions, and proceeds via linearization of the concave part.

To use CCCP for our problem we observe that from the discussion of the previous section it follows that our objective can indeed be decomposed into the following convex and concave components:

$$h_\cup(w) = \ell(w) + \beta f_1(\pi(w)), \quad h_\cap(w) = -\beta f_2(\pi(w)),$$

where $\cap$ and $\cup$ correspond to the convex and concave parts, respectively. To linearize $h_\cap(w)$ around a point $w^t$, we need to find its subgradient at that point. The next proposition, which follows easily from Hazan and Kale [2009], gives the subgradient of $f_2(\pi(w))$:

**Proposition 4.3.** *The subgradient of $f_2(\pi(w))$ is given by the vector $v$ defined as follows.[5] The coordinates in $v$ corresponding to $w_{ij}$ are given by:*

$$v_{ij} = \begin{cases} sign(w_{ij}) & ij \in T(\pi(w)) \\ 0 & otherwise \end{cases}$$

*where sign is taken element wise. The other coordinates of $v$ (corresponding to $w_i$) are zero.*

We can now specify the resulting algorithm, which we call CRANK for circuit-rank regularizer.

---
**Algorithm 1** The CRANK algorithm
---
**Input:** $w^1, \beta$
**for** $t = 1, \ldots$ **do**
 $h^t(w) = \ell(w) + \beta\|w\|_1 - \beta v(w^t)^\top w$
 $w^{t+1} = \operatorname{argmin}_w h^t(w)$
**end for**
---

The objective $h^t(w)$ to be minimized at each iteration is a convex function, which can be optimized using any convex optimization method. In this work we use the stochastic Frank-Wolfe procedure recently proposed by Lacoste-Julien et al. [2013].[6] The advantage of this approach is that the updates are simple, and it generates primal and dual bounds which help monitor convergence. In practice, we do not solve the inner optimization problems exactly, but rather up to some primal-dual gap.

---
[4]To be precise, we are using the more general DC programming framework [Tao and An, 1997], which can be applied to non differentiable functions.

[5]In cases where several $w_{ij}$ are equal, there are multiple subgradients.

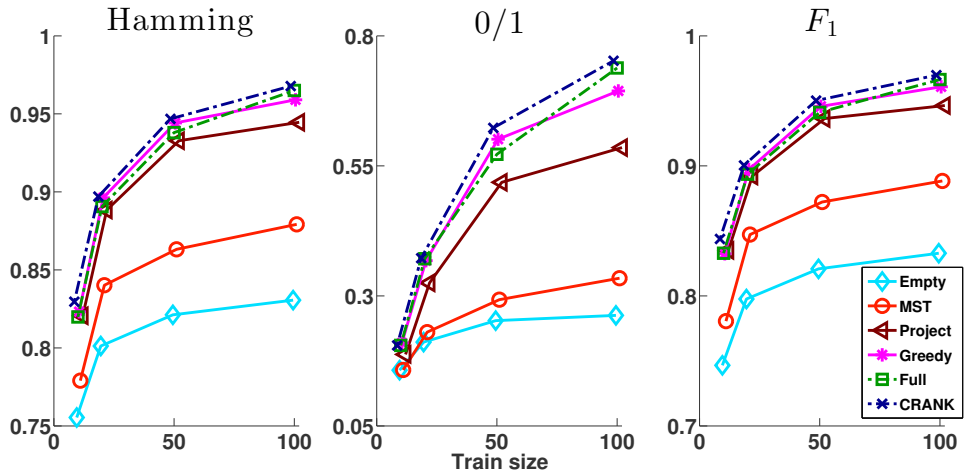

Figure 1: Average test performance as a function of the number of training samples for the synthetic datasets.

## 5 Experiments

In this section we evaluate the proposed algorithm on multi-label classification tasks and compare its performance to several baselines. In this task the goal is to predict the subset of labels which best fits a given input. We use the model presented in Finley and Joachims [2008], where each possible label $y_i \in \{0, 1\}$ is associated with a weight vector $w_i$, the singleton scores are given by $w_i^\top x y_i$, and the pairwise scores are simply $w_{ij} y_i y_j$ (i.e., $w_{ij}$ is scalar).

We compare our **CRANK** algorithm to the following baselines: The **Empty** model learns each label prediction independently of the other labels; The **Full** model learns a model that can use all pairwise dependencies; The **Greedy** trainer starts with the empty model and at each iteration adds the edge which achieves the largest gain in objective while not forming a cycle, until no more edges can be added; The **Project** algorithm, runs CRANK starting from the weights learned by the Full algorithm, and using a large penalty $\beta$;[7] The final baseline is an **MST** algorithm, which calculates the gain in objective for each edge separately, takes a maximum-spanning-tree over these weights, and then re-trains the resulting tree. Since CCCP may be sensitive to its starting point, we restart CRANK from 10 random points and choose the one with lowest objective (we run those in parallel). We apply the stochastic Frank-Wolfe algorithm [Lacoste-Julien et al., 2013] to optimize the weights in all algorithms. The Full and CRANK algorithms operate on non-tree graphs, and thus use an LP relaxation within the training loss (see Section 4.1). Since the model trained with Full is not tree structured, we also needs to use LP relaxation at test time (see implications on runtime below). We used the GLPK solver for solving the LPs.

---

[6]We modified the algorithm to handle the $\ell_1 + \ell_2$ case.
[7]The Project scheme thus trains a model consisting of the maximum-spanning-tree over the weights learned by Full, and can be viewed as a "tree-projection" of the full model.

To measure the performance of the algorithms, we consider three accuracy measures: **Hamming**, **zero-one**, and $\boldsymbol{F_1}$ (averaged over samples). See [Zhang and Schneider, 2012, Dembczynski et al., 2010] for similar evaluation schemes. Regularization coefficients were chosen using cross-validation. The parameter $\beta$ in CRANK was gradually increased until a tree structure was obtained.

**Synthetic Data:** We first show results for synthetic data where $x \in \mathbb{R}^4$. The data was created as follows: a random tree $T$ over $n = 10$ variables was picked and corresponding weights $w \in \mathcal{W}_T$ were sampled. Train and test sets were generated randomly and labeled using $w$. Test set size was 1000 and train set size varied. Results (averaged over 10 repetitions) are shown in Figure 1.

We observe that the structured models do significantly better than the Empty model. Additionally, we see that the Full, Greedy, and CRANK algorithms are comparable in terms of prediction quality, with a slight advantage for CRANK over the others. We also notice that Project and MST do much worse than the other structured models.

|        | Hamming    | 0/1        | $F_1$     | Hamming    | 0/1        | $F_1$     |
|--------|------------|------------|-----------|------------|------------|-----------|
|        | Scene      |            |           | Emotions   |            |           |
| CRANK  | 90.5 (2)   | 58.9 (2)   | 64.8 (2)  | **79.2** (1) | 29.2 (2) | 60.5 (2)  |
| Full   | **90.7** (1) | **62.2** (1) | **67.8** (1) | 79.0 (3) | **33.7** (1) | **62.5** (1) |
| Greedy | 90.2 (3)   | 56.9 (3)   | 62.6 (3)  | 78.5 (4)   | 24.3 (4)   | 54.5 (4)  |
| Project| 89.5 (5)   | 52.1 (5)   | 59.2 (5)  | 77.6 (5)   | 20.8 (5)   | 49.8 (5)  |
| MST    | 89.9 (4)   | 53.0 (4)   | 59.6 (4)  | 79.1 (2)   | 28.2 (3)   | 57.5 (3)  |
| Empty  | 89.3 (6)   | 49.5 (6)   | 56.5 (6)  | 76.7 (6)   | 20.3 (6)   | 48.5 (6)  |
|        | Medical    |            |           | Yeast      |            |           |
| CRANK  | **96.9** (1) | 74.0 (2) | 78.2 (3)  | 80.1 (2)   | 17.6 (2)   | 60.4 (3)  |
| Full   | **96.9** (1) | **75.0** (1) | **78.3** (1) | **80.2** (1) | **19.0** (1) | **60.9** (1) |
| Greedy | **96.9** (1) | 74.0 (2) | **78.3** (1) | NA       | NA         | NA        |
| Project| 96.7 (4)   | 72.7 (4)   | 77.0 (5)  | 80.1 (2)   | 16.4 (3)   | 60.7 (2)  |
| MST    | 96.7 (4)   | 71.9 (5)   | 77.5 (4)  | 80.1 (2)   | 16.1 (4)   | 60.3 (4)  |
| Empty  | 96.4 (6)   | 71.0 (6)   | 76.1 (6)  | 79.8 (5)   | 12.1 (5)   | 58.0 (5)  |

Table 1: Performance on test data for real-world multi-label datasets. The rank of each algorithm for each dataset and evaluation measure is shown in brackets. Greedy was too slow to run on Yeast.

**Real Data:** We next performed experiments on four real-world datasets.[8] In the **Scene** dataset the task is to classify a given image into several outdoor scene types (6 labels, 294 features). In the **Emotions** dataset we wish to assign emotional categories to musical tracks (6 labels, 72 features). In the **Medical** dataset the task is document classification (reduced to 10 labels, 1449 features). This is the experimental setting used by Zhang and Schneider [2012]. Finally, to test how training time scales with the number of labels, we also experiment with the **Yeast** dataset, where the goal is to predict which functional classes each gene belongs to (14 labels, 103 features). The results are summarized in Table 1.

We first observe that in this setting the Full model generally has the best performance and Empty the worst. As before, CRANK is typically close to Full and outperforms Greedy and the other baselines in the majority of the cases.

**Runtime analysis:** In Table 2 we report train and test run times, relative to the Full model, for the Yeast dataset.

Table 2: Running times for the Yeast dataset.

| Time  | CRANK | Full | Project | MST  | Empty |
|-------|-------|------|---------|------|-------|
| Train | 1.82  | 1.0  | 0.17    | 1.32 | 0.01  |
| Test  | 0.02  | 1.0  | 0.06    | 0.02 | 0.02  |

It is important to note that the greedy algorithm is very slow to train, since it requires solving $O(n^3)$ training problems. It is thus impractical for large problems, and this is true even for the Yeast dataset which has only $n = 14$ labels (and hence does not appear in Table 2). The Full model on the other hand has much longer test times (compared to the tree-based models), since it must use an LP solver for prediction. CRANK has a training time that is reasonable (comparable to Full) and a much better test time. On the Yeast dataset prediction with CRANK is 50 times faster than with Full.

---

[8]Taken from Mulan (http://mulan.sourceforge.net)

In conclusion, CRANK seems to strike the best balance in terms of accuracy, training time, and test time: it achieves an accuracy that is close to the best among the baselines, with a scalable training time, and very fast test time. This is particularly appealing for applications where we can afford to spend some more time on training while test time is critical (e.g., real-time systems).

## 6 Related Work

The problem of structure learning in graphical models has a long history, dating back to the celebrated algorithm by Chow and Liu [1968] for finding a maximum likelihood tree in a generative model. More recently, several elegant works have shown the utility of $\ell_1$ regularization for structure learning in generative models [Friedman et al., 2008, Lee et al., 2007, Ravikumar et al., 2008, 2010]. These works involve various forms of $\ell_1$ regularization on the model parameters, coupled with approximation of the likelihood function (e.g., pseudo-likelihood) when the underlying model is non-Gaussian. Some of these works also provide finite sample bounds on the number of of samples required to correctly reconstruct the model structure. Unlike ours, these works focus on generative models, a difference that can be quite fundamental. For example, as we proved in Section 3, learning trees is a problem that is NP-hard in the $M^3N$ setting while it is *polynomial* in the number of variables in the generative one. We note that we are not aware of finite sample results in the discriminative setting, where a different MRF is considered for each input $x$.

In the discriminative setting, Zhu et al. [2009] define an $\ell_1$ regularized $M^3N$ objective and present algorithms for its optimization. However, this approach does not consider the structure of the underlying graphical model over outputs $y$. Torralba et al. [2004] propose a greedy procedure based on boosting to learn the structure of a CRF, while Schmidt et al. [2008] present a block-$\ell_1$ regularized pseudo-likelihood approach for the same task. While these methods do consider the structure of the graphical model, they do not attempt to produce tractable predictors. Further, they do not aim to learn models using a max-margin objective.

Bradley and Guestrin [2010] address the problem of learning trees in the context of conditional random fields. They show that using a particular type of tractable edge scores together with a maximum-spanning-tree (MST) algorithm may fail to find the optimal tree structure. Our work differs from theirs in several aspects. First, we consider the max-margin setting for structured prediction rather than the CRF setting. Second, they assume that the feature function $\phi(x, y)$ has a particular form where $y$ and $x$ are of the same order and where the outputs depend only on local inputs. Finally, we do not restrict our attention to MST algorithms. Consequently, our hardness result is more general and the approximations we propose are quite different from theirs. Finally, Chechetka and Guestrin [2010] also consider the problem of learning tree CRFs. However, in contrast to our work, they allow the structure of the tree to depend on the input $x$, which is somewhat more involved as it requires learning an additional mapping from inputs to trees.

## 7 Conclusion

We tackled the challenge of learning tree structured prediction models. To the best of our knowledge, ours is the first work that addresses the problem of structure learning in a discriminative $M^3N$ setting. Moreover, unlike common structured sparsity approaches that are used in the setting of generative and conditional random field models, we explicitly target tree structures due to their appealing properties at test time. Following a proof that the task is NP-hard in general, we proposed a novel approximation scheme that relies on a circuit rank regularization objective that penalizes non-tree models, and that can be optimized using the CCCP algorithm. We demonstrated the effectiveness of our CRANK algorithm in several real-life domains. Specifically, we showed that CRANK obtained accuracies very close to those achieved by a full-dependence model, but with a much faster test time.

Many intriguing questions arise from our work: Under which conditions can we learn the optimal model, or guarantee approximation quality? Can we extend our framework to the case where the tree depends on the input? Can we use a similar approach to learn other graphs, such as low treewidth or high girth graphs? Such settings introduce novel forms of *graph-structured* sparsity which would be interesting to explore.

## Acknowledgments

This research is funded by the ISF Centers of Excellence grant 1789/11, and by the Intel Collaborative Research Institute for Computational Intelligence (ICRI-CI). Ofer Meshi is a recipient of the Google Europe Fellowship in Machine Learning, and this research is supported in part by this Google Fellowship.

## A  Realizing the parameters

Here we complete the hardness proof in Section 3 by describing the training set that realizes the parameters of Eq. (8) and Eq. (9). The approach below makes use of two training samples to constrain each parameter, one to upper bound it and one to lower bound it. Appealingly, the training samples are constructed in such a way that other samples do not constrain the parameter value, allowing us to realize the needed value for each and every parameter in the model.

The construction is similar to Sontag et al. [2010]. For all parameters it consists of the following steps: (i) define an assignment $x(y)$; (ii) identify two $y$ values that potentially maximize the score; and (iii) show that these two complete assignments to $x$ and $y$ force the desired parameter value.

**Preliminaries:** Recall that the parameter vector $\theta$ is defined in Eq. (6) via a product of the features $\phi_i(y_i, x)$ and $\phi_{ij}(y_i, y_j, x)$ and the weights $w_i$ and $w_{ij}$ which do not depend on $x$ or $y$ and are shared across the parameters. For the hardness reduction, it will be convenient to set the features to indicator functions and show that the implied weight values are realizable. Specifically, we set the features to:

$$\phi_{ij}^{\text{off-diag}}(y_i, y_j) = \mathbb{1}\{y_i \neq y_j\} \ \forall i, j$$
$$\phi_{ij}^{\text{diag}}(y_i, y_j) = \mathbb{1}\{y_i = y_j = i \text{ or } y_i = y_j = j\} \ \forall i, j$$
$$\phi_1^{\text{bound}}(y_1) = \mathbb{1}\{y_1 = 0\}$$

Recall that to realize the desired parameters $\theta$, we need to introduce training samples such that for all $i, j$:

$$w_{ij}^{\text{off-diag}} = -n^2, \quad w_{ij}^{\text{diag}} = 1, \quad w_1^{\text{bound}} = D,$$

with the dimension of $w$ equalling $2|E(G)| + 1$.

Finally, using $N_i$ to denote the set of neighbors of node $i$ in the graph $G$, we will use the following (large) value to force variables not to take some chosen states:

$$\gamma_i = \begin{cases} 1 + |N_i|(n^2+1) & i \neq 1 \\ 1 + |N_i|(n^2+1) + D & i = 1 \end{cases}$$

**Realizing the weight $w_1^{\text{bound}}$:** We define $x(y)$ as:

$$x_1(y_1) = \begin{cases} 0 & y_1 = 0 \\ -D & y_1 = 1 \\ -\gamma_1 & y_1 \geq 2 \end{cases}$$

$$x_i(y_i) = \begin{cases} 0 & y_i = 2 \\ -\gamma_i & y_i \neq 2 \end{cases} \quad \text{for all } i \neq 1$$

Recalling the definition of $\gamma$ above, this implies that the only assignments to $y$ that can maximize the score of Eq. (6) are $(0, 2, 2, ..., 2)$ and $(1, 2, 2, ..., 2)$. In particular, we have:

$$S(0, 2, 2, ..., 2; x, w) = \sum_{k \in N_1} w_{1k}^{\text{off-diag}} + x_1(0) + \bar{S}$$

$$S(1, 2, 2, ..., 2; x, w) = w_1^{\text{bound}} + \sum_{k \in N_1} w_{1k}^{\text{off-diag}} + x_1(1) + \bar{S}$$

where $\bar{S}$ is the sum of all components that do not involve the first variable.

For the final step we recall that the weights $w$ have to satisfy the constraints: $S(y^m; x^m, w) \geq S(y; x^m, w)$ for all $m, y$. Thus, we will define two instances $(x^m, y^m)$ for which some $y$ assignment will constrain the weight as needed (in both cases, $x^m$ is defined as above). When $y^{(m)} = (0, 2, 2, \ldots, 2)$, the assignment $y = (1, 2, 2, \ldots, 2)$ yields $w_1^{\text{bound}} \leq D$ and all other assignments do no further constrain the weight. Similarly, for $y^{(m')} = (1, 2, 2, \ldots, 2)$, the assignment $y = (0, 2, 2, \ldots, 2)$ yields $w_1^{\text{bound}} \geq D$. Together, the two assignments constrain the weight parameter to $w_1^{\text{bound}} = D$, as desired.

**Realizing the weights $w_{ij}^{\text{off-diag}}$:** We define $x(y)$ as:

$$x_i(y_i) = \begin{cases} 0 & y_i = 0 \\ -\gamma_i & y_i \neq 0 \end{cases}, \quad x_j(y_j) = \begin{cases} 0 & y_j = 0 \\ n^2 & y_j = 1 \\ -\gamma_j & y_j \geq 2 \end{cases}$$

$$x_k(y_k) = \begin{cases} 0 & y_k = k \\ -\gamma_k & y_k \neq k \end{cases} \quad \text{for all } k \neq i, j$$

This implies that except for $i$ and $j$, all $y_k$'s must take their corresponding assignment so that $y_k = k$. W.l.g., suppose that $i = 1$ and $j = 2$. The only assignments that can maximize the score are $(0, 0, 3, 4, 5, ..., n)$ and $(0, 1, 3, 4, 5, ..., n)$ with values:

$$S(0, 0, 3, 4, 5, ..., n; x, w) =$$
$$\sum_{k \in N_i} w_{ik}^{\text{off-diag}} + \sum_{k' \in N_j} w_{jk'}^{\text{off-diag}} + x_i(0) + x_j(0) + \bar{S}$$

$$S(0, 1, 3, 4, 5, ..., n; x, w) =$$
$$w_{ij}^{\text{off-diag}} + \sum_{k \in N_i} w_{ik}^{\text{off-diag}} + \sum_{k' \in N_j} w_{jk'}^{\text{off-diag}} + x_i(0) + x_j(1) + \bar{S}$$

As before, setting $y^{(m)} = (0, 0, 3, 4, 5, ..., n)$ and then $y^{(m')} = (0, 1, 3, 4, 5, ..., n)$ yields the constraint $w_{ij}^{\text{off-diag}} = -n^2$.

**Realizing the weights $w_{ij}^{\text{diag}}$:** We define $x(y)$ as:

$$x_i(y_i) = \begin{cases} 0 & y_i = i \\ -\gamma_i & y_i \neq i \end{cases}, \quad x_j(y_j) = \begin{cases} n^2 & y_j = 0 \\ -1 & y_j = i \\ -\gamma_j & y_j \notin \{0, i\} \end{cases}$$

$$x_k(y_k) = \begin{cases} 0 & y_k = k \\ -\gamma_k & y_k \neq k \end{cases} \quad \text{for all } k \neq i, j$$

As before, for all $k \neq i, j$ the assignment is forced to $y_k = k$. The maximizing assignments are now $(1, 0, 3, 4, 5, ..., n)$ and $(1, 1, 3, 4, 5, ..., n)$ (assuming w.l.g., $i = 1, j = 2$) with score values:

$$S(1, 0, 3, 4, 5, ..., n; x, w) =$$
$$w_{ij}^{\text{off-diag}} + \sum_{k \in N_i} w_{ik}^{\text{off-diag}} + \sum_{l \in N_j} w_{jl}^{\text{off-diag}} + x_i(i) + x_j(0) + \bar{S}$$

$$S(1, 1, 3, 4, 5, ..., n; x, w) =$$
$$w_{ij}^{\text{diag}} + \sum_{k \in N_i} w_{ik}^{\text{off-diag}} + \sum_{l \in N_j} w_{jl}^{\text{off-diag}} + x_i(i) + x_j(i) + \bar{S}$$

Now, setting $y^{(m)} = (1, 0, 3, 4, 5, ..., n)$, the assignment $y = (1, 1, 3, 4, 5, ..., n)$ implies $w_{ij}^{\text{off-diag}} + n^2 \geq w_{ij}^{\text{diag}} - 1$. Since we already have that $w_{ij}^{\text{off-diag}} = -n^2$, we obtain $w_{ij}^{\text{diag}} \leq 1$. Similarly, adding $y^{(m')} = (1, 1, 3, 4, 5, ..., n)$ implies $w_{ij}^{\text{diag}} \geq 1$, so together we have $w_{ij}^{\text{diag}} = 1$, as required.

Importantly, our trainset realizes the same edge parameters for any possible tree, since edge weights are constrained independently of other edges.